\documentclass[10pt,twocolumn,letterpaper]{article}

\usepackage{cvpr}              

\usepackage{multirow}  		
\usepackage{amsmath} 		
\usepackage{wrapfig} 		
\usepackage{graphicx} 		
\usepackage{subcaption} 	
\usepackage{mathrsfs}
\usepackage{bm}
\usepackage{algorithm}
\usepackage[noend]{algorithmic} %
\usepackage{colortbl}
\usepackage[accsupp]{axessibility}  
\definecolor{cvprblue}{rgb}{0.21,0.49,0.74}
\usepackage[pagebackref,breaklinks,colorlinks,allcolors=cvprblue]{hyperref}

\usepackage{comment}

\newcommand{\ul}[1]{\underline{#1}}

\def\paperID{***} 
\def\confName{CVPR}
\def\confYear{2026}

\title{Few-Shot Incremental 3D Object Detection in Dynamic Indoor Environments}

\author{
Yun Zhu\textsuperscript{\rm 1}\thanks{This work was done during Yun Zhu’s visit to the IMPL Lab at SUTD.}, 
~Jianjun Qian\textsuperscript{\rm 1},
~Jian Yang\textsuperscript{\rm 1}, 
Jin Xie\textsuperscript{\rm 2}\thanks{Corresponding author},  
Na Zhao\textsuperscript{\rm 3}\footnotemark[2]\\
\textsuperscript{\rm 1}Nanjing University of Science and Technology\\
\textsuperscript{\rm 2}Nanjing University,
\textsuperscript{\rm 3}Singapore University of Technology and Design \\
\tt\small \{zhu.yun, csjqian, csjyang\}@njust.edu.cn; csjxie@nju.edu.cn; na\_zhao@sutd.edu.sg
}

\begin{document}
\maketitle
\begin{abstract}
Incremental 3D object perception is a critical step toward embodied intelligence in dynamic indoor environments. 
However, existing incremental 3D detection methods rely on extensive annotations of novel classes for satisfactory performance. To address this limitation, we propose \textbf{FI3Det}, a \underline{\textbf{F}}ew-shot \underline{\textbf{I}}ncremental \underline{\textbf{3D}} \underline{\textbf{Det}}ection framework that enables efficient 3D perception with only a few novel samples by leveraging vision-language models (VLMs) to learn knowledge of unseen categories. FI3Det introduces a VLM-guided unknown object learning module in the base stage to enhance perception of unseen categories. Specifically, it employs VLMs to mine unknown objects and extract comprehensive representations, including 2D semantic features and class-agnostic 3D bounding boxes. To mitigate noise in these representations, a weighting mechanism is further designed to re-weight the contributions of point- and box-level features based on their spatial locations and feature consistency within each box. Moreover, FI3Det proposes a gated multimodal prototype imprinting module, where category prototypes are constructed from aligned 2D semantic and 3D geometric features to compute classification scores, which are then fused via a multimodal gating mechanism for novel object detection. As the first framework for few-shot incremental 3D object detection, we establish  both batch and sequential evaluation settings on two datasets, ScanNet V2 and SUN RGB-D, where FI3Det achieves strong and consistent improvements over baseline methods. Code is available at \emph{\url{https://github.com/zyrant/FI3Det}}.
\end{abstract}    
\definecolor{light green}{RGB}{0,224,0}
\definecolor{light orange}{RGB}{255,175,0}

\section{Introduction}
\label{sec:intro}
Real-world indoor embodied environments are \textit{dynamic}, where new object categories continuously emerge over time. However, most existing 3D object detection methods~\cite{cagroup, voxelnext, fcaf3d, tr3d, voxelrcnn, uncertainty, dual, wu2026ccf} are based on a \textit{static} paradigm, assuming that annotations for all categories are available within a single session, which limits their applicability in real-world 3D applications.
\begin{figure}[t]
	\centering
	\includegraphics[width=1.0\linewidth]{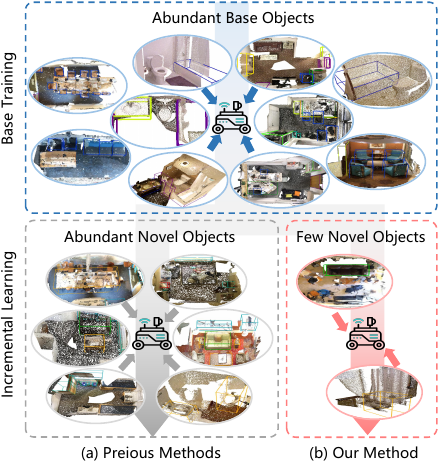}
	\vspace{-15pt}
	\caption{
		Comparison between incremental 3D object detection (In3Det) and our FI3Det setting. (a) In3Det methods assume abundant annotated samples for novel classes, limiting their applicability in dynamic embodied environments. (b) In contrast, FI3Det requires only a few annotated novel samples, enabling efficient and scalable incremental perception.
	}
	\label{fig:fsi3od}
	\vspace{-15pt}
\end{figure}
To address this limitation, recent research has begun exploring how to enable models to incrementally recognize previously unseen categories in indoor embodied environments. 

As a pioneering effort, SDCoT~\cite{sdcot} introduces the task of \textit{incremental 3D object detection}, addressing it through dual-teacher pseudo-labeling and knowledge distillation. Its extension, SDCoT++~\cite{sdcot++}, incorporates adaptive class probability calibration to enhance distillation efficiency. More recently, AIC3DOD~\cite{aic3dod} improves feature representation by optimizing incremental learning steps within a transformer architecture. Despite these advances, existing incremental methods still depend on abundant labeled samples for each new class in 3D object detection (Fig.~\ref{fig:fsi3od}a).

In contrast, the \textit{few-shot incremental 3D object detection} setting aims to progressively expand the detected category space using only a few samples per new class (Fig.~\ref{fig:fsi3od}b). This setting closely mirrors human learning, where new concepts can be acquired from limited examples while retaining prior knowledge. However, this promising direction remains largely unexplored and presents a fundamental challenge: balancing the learning of novel categories with the preservation of previously acquired knowledge under extremely limited supervision.

\begin{figure}[t]
	\centering
	\includegraphics[width=1.0\linewidth]{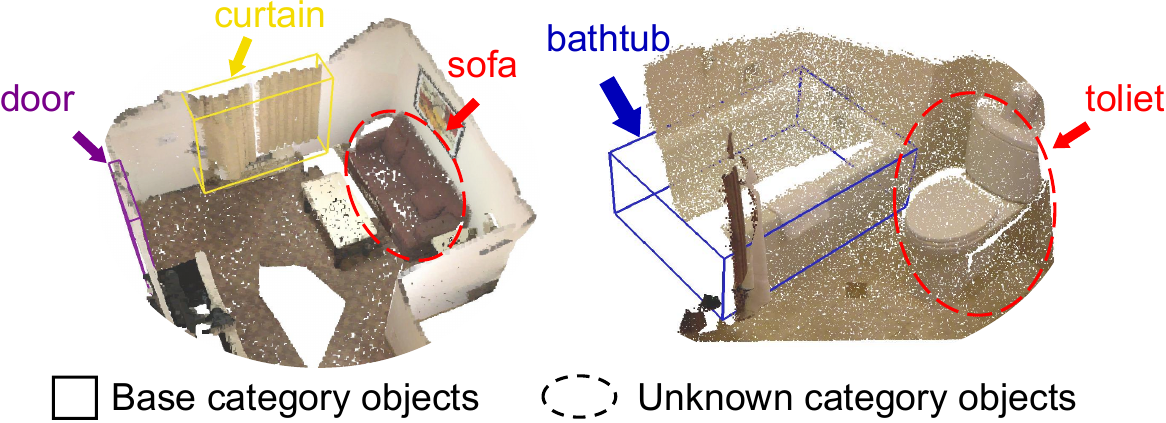}
	\vspace{-15pt}
	\caption{Correlation between base and novel category objects. In indoor 3D scenes, novel category objects tend to appear alongside base category objects. }
	\label{fig:uknowexit}
	\vspace{-10pt}
\end{figure}

To bridge this gap, we propose \textbf{FI3Det}, a few-shot incremental 3D object detection model that efficiently adapts to novel categories with only a few samples while preserving detection performance on base classes. FI3Det introduces two key components: 1) a VLM-guided unknown object learning module and 2) a gated multimodal prototype imprinting module, where the former operates during the base stage and the latter during the incremental stage.

In the base stage, we observe that novel objects already appear in the scenes, albeit without annotations, as illustrated in Fig.~\ref{fig:uknowexit}. These objects are present but undefined, and thus treated as unknown objects in the base stage. This observation motivates our VLM-guided unknown object learning module, designed to establish early awareness of novel categories during base training. Specifically, we use Vision-Language Models (VLMs)~\cite{grounding, efficientsam, sam} to generate 2D instance features and class-agnostic pseudo 3D bounding boxes, which serve as auxiliary supervision signals for unknown object learning. To mitigate noise in these representations, we introduce a weighting strategy that leverages both spatial priors of points and semantic consistency within a box to adjust the contribution of each point and box during supervision. At the point level, a Gaussian-based spatial weighting emphasizes points near the box center, as they are less affected by segmentation errors. At the box level, samples with higher feature consistency are assigned larger weights.

In the incremental stage, FI3Det introduces the gated multimodal prototype imprinting module, which follows a prototype-based incremental learning strategy to enable rapid adaptation to novel categories while preserving the decision boundaries of base classes. Based on the aligned 2D semantic features and 3D geometric features, we construct modality-specific category prototypes that jointly capture semantic and geometric cues. Building upon these prototypes, we further design a multimodal gating mechanism that adaptively fuses prototype-based classification scores across modalities for improved robustness and generalization. Our contributions are summarized as follows:
\begin{itemize}
	\item We present the first study on few-shot incremental 3D object detection in dynamic indoor environments and propose a novel framework, FI3Det, that efficiently adapts to novel objects while preserving detection performance on previously learned classes.
	\item We design a VLM-guided unknown object learning module that leverages VLM priors to learn from potential novel objects during base training, and a gated multimodal prototype imprinting module that fuses 2D semantic and 3D geometric cues for incremental learning.  
	\item Our FI3Det achieves state-of-the-art performance in both few-shot batch and sequential incremental 3D object detection on ScanNet V2~\cite{scannet} and SUN RGB-D~\cite{sunrgbd}, delivering an average 17.37$\%$ improvement on novel classes over baseline models. This demonstrates  strong ability of our framework to adapt to new categories with limited novel samples.
\end{itemize}
 \section{Related Work}
\label{sec:relatedwork}

\noindent\textbf{3D Point Cloud Object Detection.} 
As a fundamental perception in embodied intelligence tasks~\cite{jiang2023se, jiang2021sampling, Hui_2023_ICCV, hui2022learning, Wu_2025_CVPR, Xie_2025, MonoSE3, affordBot, pan2025ial, text2lidar, augrefer}, 3D point cloud understanding has achieved remarkable progress in recent years, with point-based~\cite{pointnet, pointnet++} and voxel-based~\cite{minkowski, spconv} representations. 
3D object detection representative methods such as VoteNet~\cite{votenet}, FCAF3D~\cite{fcaf3d}, and VoxelNext~\cite{voxelnext} significantly improve detection performance through effective feature aggregation and anchor-free design. Recent works like TR3D~\cite{tr3d}, SPGroup3D~\cite{spgroup3d}, VDETR~\cite{vdetr}, and OneDet3D~\cite{onenet3d} further enhance structure simplicity, instance consistency, and cross-domain generalization. However, these approaches remain limited to closed-set scenarios and struggle to generalize to novel categories in dynamic 3D environments.

\begin{figure*}[htbp]
	\centering
	\includegraphics[width=1.0\textwidth]{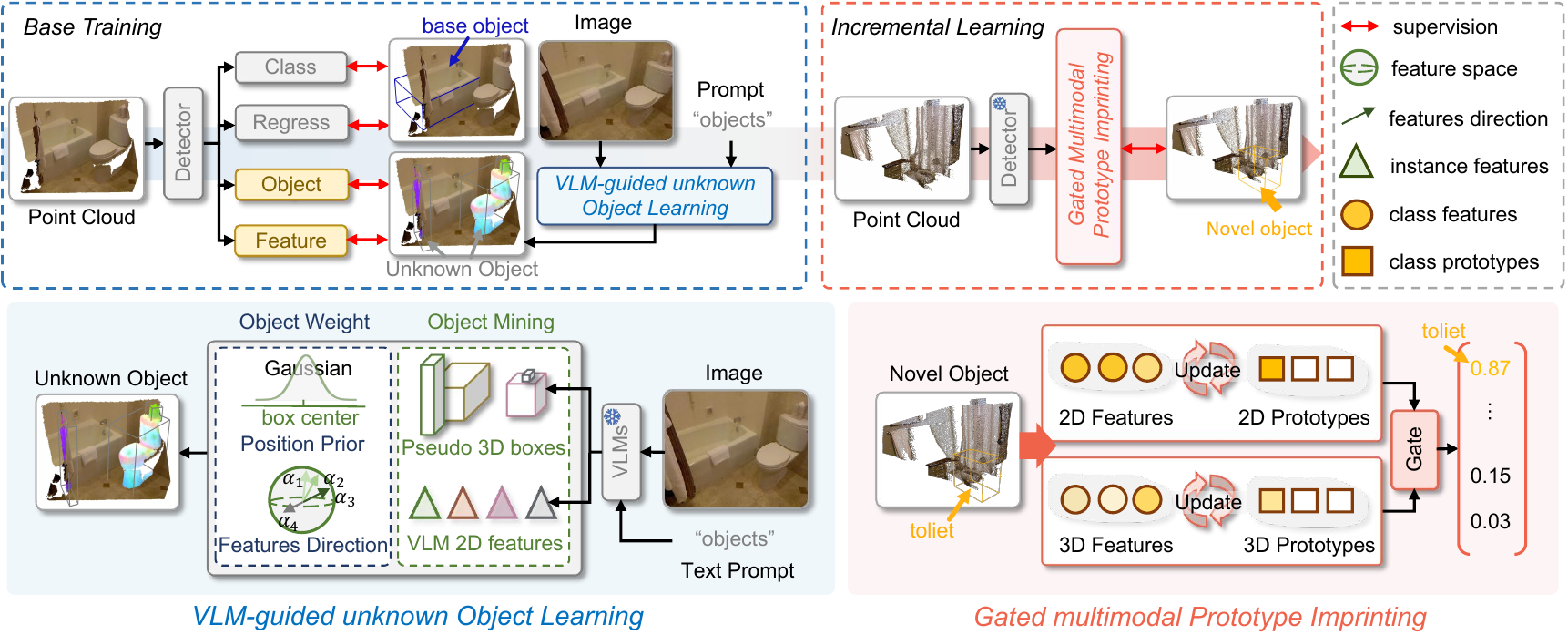}
	\caption{\textbf{Overview of our few-shot incremental 3D object detection model}. The model consists of two parts: base training and incremental learning. In the base stage, we introduce a VLM-guided unknown object learning module that uses 2D VLMs to generate unknown objects, thereby improving the perception of unknown objects. In the incremental stage, we propose a gated multimodal prototype imprinting module that builds 2D semantic and 3D geometric prototypes for efficient adaptation to novel categories.}
	\label{fig:framework}
	\vspace{-10pt}
\end{figure*}

\noindent\textbf{Incremental 
Object Detection.} 
Dynamic indoor environments pose significant challenges to 3D object detection, requiring continuous perception and adaptation. Incremental 3D detection~\cite{sdcot, sdcot++, aic3dod, continual} addresses base forgetting through pseudo-labeling~\cite{sdcot++} or layout learning~\cite{aic3dod}, but needs extensive novel class objects. Few-shot incremental learning bridges this gap by adapting to new classes efficiently under limited data without forgetting.
In 2D domain~\cite{once, sylph, incre_detr, chen2025enhancing, jiang2025revisiting, zhao2025attraction, liu2025sec}, methods such as~\cite{chen2025enhancing, jiang2025revisiting, zhao2025attraction, liu2025sec} have made contributions to few-shot incremental image classification. While methods such as ONCE~\cite{once}, Sylph~\cite{sylph}, and the recent IL-DETR~\cite{incre_detr} demonstrate effective adaptation in 2D detection, the 3D domain remains under-explored. The complex layouts and diverse object compositions of 3D indoor scenes further amplify inter-class variations, making few-shot incremental 3D detection particularly difficult. In this paper, we propose a novel few-shot incremental 3D object detection model that enables adaptation to new categories with limited data.

\noindent\textbf{Data-efficient 3D Object Detection.}
Recent advances in data-efficient 3D object detection~\cite{backtoreal, sess, ss3d, cpdet3d, diffusion, ioumatch, dual, dqs3d, weakly3d, wang2024syn} have achieved impressive results in closed-set settings.
Building upon vision-language models (VLMs)~\cite{grounding, efficientsam, sam, yolo_world}, recent works~\cite{generalized, mixsup, general, sp3d} such as GFS-VL~\cite{generalized}, MixSup~\cite{mixsup}, and SP3D~\cite{sp3d} leverage VLM-generated pseudo labels to enhance data efficiency, yet they mainly emphasize pseudo label generation and neglect feature-level learning.
In contrast, few-shot approaches like Prototypical VoteNet~\cite{prototypical} and its variant~\cite{vae} enable novel class detection through prototype interaction but ignore base classes. To address these limitations, we propose a novel approach that jointly learns box- and feature-level representations, enabling the model to adapt to novel categories while preserving knowledge of base classes.
    
\section{Method}
\label{sec:method}

We define the few-shot incremental 3D object detection task as training a parametric model on a sequence of sessions.  The first session, referred to as the \textit{base session}, contains a set of base object categories $\mathcal{C}_{\text{base}}$ with abundant annotated samples $(\bm{\mathcal{X}}^{0}, \bm{\mathcal{Y}}^{0})$, allowing the model to learn fundamental 3D object representations. Subsequent sessions $t \in \{1, \dots, T\}$ are \textit{incremental sessions}, each introducing a small set of novel categories $\mathcal{C}_{\text{novel}}^{(t)}$ and a corresponding data distribution $(\bm{\mathcal{X}}^{(t)}, \bm{\mathcal{Y}}^{(t)})$ with only a few annotated samples. Different sessions contain disjoint class sets, \textit{i.e.}, $\mathcal{C}_{\text{novel}}^{(m)} \neq \mathcal{C}_{\text{novel}}^{(n)}$ for $m \neq n$. The cumulative category space after session $t$ is defined as
\begin{equation}
	\mathcal{C}_{\text{all}}^{(t)} = 
	\mathcal{C}_{\text{base}} \cup 
	\mathcal{C}_{\text{novel}}^{(1)} \cup 
	\dots \cup 
	\mathcal{C}_{\text{novel}}^{(t)}.
\end{equation}
The goal of this task is to continuously learn novel categories from limited samples while preserving prior knowledge without accessing previous data. 
After $T$ incremental sessions, the trained model should be capable of detecting all objects in $\mathcal{C}_{\text{all}}^{(T)}$.

To 
address this task, we propose an innovative framework designed to learn 
novel object categories from only a few samples. An overview of the framework is illustrated in Fig.~\ref{fig:framework}, including unknown object learning module (Sec.~\ref{sec:lol}) and multimodal prototype imprinting module (Sec.~\ref{sec:mci}). The overall learning strategy is presented in Sec.~\ref{sec:ts}.

\subsection{VLM-guided Unknown Object Learning} 
\label{sec:lol}
In indoor scenarios, it is often observed that novel objects naturally appear in training scenes even during the base class stage, offering valuable cues about unseen categories. 
Motivated by this observation, we propose a VLM-guided unknown object learning module 
that leverages the zero-shot recognition capability of VLMs~\cite{grounding, efficientsam}. This module provides auxiliary supervision for unknown objects during base class training and consists of two components: (1) unknown object mining, which employs VLMs to generate comprehensive representations for unknown objects, and (2) an unknown object weighting module, which adaptively re-weights these representations to suppress noise.

\noindent\textbf{Unknown Object Mining.} 
The most straightforward way to utilize VLMs is to adopt their class-agnostic pseudo 3D boxes as supervision. However, box-level cues provide only localization and offer no semantic perception. Hence, we introduce unknown object mining that employs both pseudo 3D boxes and VLM-derived 2D features to achieve comprehensive perception of unknown objects. Specifically, given $i$-th 3D scene $\bm{P} \in \mathbb{R}^{N \times 6}$ from the base class training set, we begin by extracting its corresponding RGB image $\bm{I} \in \mathbb{R}^{H \times W \times 3}$, and then obtain the 2D masks and visual features from it using pre-trained VLMs~\cite{grounding, efficientsam}, where $N$ is the number of 3D points, and $H$ and $W$ denote the image height and width. This step can be formulated as:
\begin{equation}
	(\bm{V}^{2D}, \bm{M}^{2D}) = 
	\psi_{\text{vlm}}(\bm{I}),
\end{equation}
where $\psi_{\text{vlm}}(\cdot)$ denotes the frozen VLMs, and $\bm{M}^{2D} \in \mathbb{R}^{H \times W \times J}$ and $\bm{V}^{2D} \in \mathbb{R}^{H \times W \times K}$ represent the 2D masks and feature maps, with $J$ masks and $K$-dimensional features, respectively. For simplicity, we omit the scene index $i$ in the above and following formulas, as all operations are performed per scene unless otherwise specified.

Next, the 2D masks are lifted into 3D space using the camera poses and depth maps, producing 3D masks $\bm{M}^{3D} \in \mathbb{R}^{N \times J}$. For $j$-th 3D mask, we obtain an instance feature $\bm{f}_{j}^{2D} \in \mathbb{R}^{K}$ 
by averaging the VLM features within the mask region, and fit a 3D bounding box $\bm{b}_{j}^{3D} \in \mathbb{R}^{7}$ over the corresponding 3D points. Ultimately, the $j$-th auxiliary supervision set is defined as:
\begin{equation}
	\bm{R}_j =
	\left\{
	(\bm{b}_{j}^{3D}, \bm{f}_{j}^{2D}) 
	\;\middle|\;
	j = 1, \ldots, J
	\right\},
\end{equation}
where each pair $(\bm{b}_{j}^{3D}, \bm{f}_{j}^{2D})$ represents a 3D object annotated by its geometric box and 2D semantic feature. To learn unknown objects, as shown in Fig. \ref{fig:framework}, we add an objectness head to identify potential unknown object regions and a feature head to align $\bm{F}^{2D} \in \mathbb{R}^{J \times K}$ with the 3D space. The resulting aligned features are represented as $\hat{\bm{F}}^{2D} \in \mathbb{R}^{N \times K}$.

\noindent\textbf{Unknown Object Weighting.} 
Although directly using mined unknown objects can provide pseudo supervision, their reliability is limited compared to human annotations. The resulting noise may hinder effective learning of unknown objects. To address this, we introduce an unknown object weighting mechanism that adaptively adjusts the contribution of pseudo boxes based on both point- and box-level confidence, ensuring more reliable supervision.

First, we construct a point-level confidence based on the relative distance between each point and the object center. 
This design is based on the intuition that segmentation results closer to the object center are usually more reliable, while farther away tend to be noisier. 
For $e$-th 3D point $\bm{p}_e \in \mathbb{R}^{3}$ belonging to the $j$-th pseudo box, its spatial weight is defined as:
\begin{equation}
	w_{e,j}^{\text{point}} = 
	\exp\!~(
	-\frac{\|\bm{p}_e - \bm{c}_j\|_2^2}{2\sigma^2}
	),
\end{equation}
where $\bm{p}_e$ and $\bm{c}_j$ denote the coordinates of the $e$-th point and the center of the $j$-th pseudo box, respectively, and $\sigma$ controls the spatial concentration around the center.

Second, we introduce a box-level weighting based on aligned feature consistency, as features belonging to the same object are expected to share coherent semantic features. Therefore, low intra-box feature consistency suggests that the corresponding pseudo 3D box is unreliable. We can formulate this step as:
\begin{equation}
	w_j^{\text{box}} =
	\|
	\frac{1}{|\mathcal{B}_j|}
	\sum_{\hat{\bm{f}}_{e}^{2D} \in \mathcal{B}_j} 
	\operatorname{norm}(\hat{\bm{f}}_{e}^{2D})
	\|_2,
\end{equation}
where $\mathcal{B}_j$ denotes the corresponding aligned features $\hat{\bm{f}}_{e}^{2D}$ within the $j$-th pseudo box, and $\operatorname{norm}(\cdot)$ denotes the normalization applied to each feature. A larger $w_j^{\text{box}}$ indicates higher semantic consistency, thus reflecting greater box reliability. Finally, the overall weight for each feature is computed as the product of the point- and box-level terms.

\begin{figure}[t]
	\centering
	\includegraphics[width=1.0\linewidth]{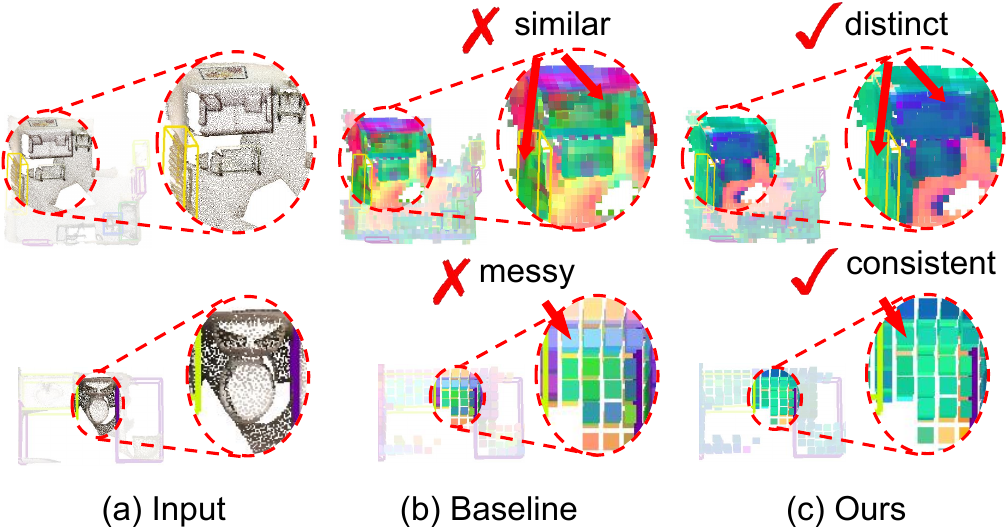}
	\caption{Visualization comparison of features. In (b), the baseline produces similar feature object across different objects and inconsistent features within the same object. In contrast, with the unknown object learning module, our method (c) generates more discriminative inter-object features and maintains intra-object consistency.}
	\label{fig:feat_show}
	\vspace{-10pt}
\end{figure}

\noindent\textbf{Visualization.}  
As illustrated in Fig.~\ref{fig:feat_show}, the proposed VLM-guided unknown object learning enables our model to learn more discriminative and consistent local features, further validating the effectiveness of the proposed module.
\subsection{Gated multimodal Prototype Imprinting}
\label{sec:mci}
Although imprinting-based prototype representations~\cite{imprinting} can preserve the decision boundaries of previously learned categories during incremental learning, directly applying them to novel classes still has two limitations. First, existing methods lack the novel object perception and localization capability provided by our unknown object learning module. Second, they typically rely on a single modality and fail to exploit the complementary strengths of 2D semantic and 3D geometric features, which restricts their generalization to new categories.
To address these issues, we propose a gated multimodal prototype imprinting mechanism consisting of: (1) modality-specific prototype updating for both 2D and 3D representation learning, and (2) multimodal score gating for adaptive cross-modal fusion.
Algorithm \ref{algo:gmproto} summarizes our gated multimodal prototype imprinting.

\noindent\textbf{Modality-specific Prototype Updating.}
With the proposed unknown object learning module, the detector trained in base training already acquires the ability to understand and localize novel class objects. Therefore, in the incremental stage, we can obtain the class representations by updating their prototypes with the extracted novel class features. To jointly exploit the complementary strengths of 2D semantic and 3D geometric information, we construct modality-specific prototypes using both the aligned 2D features $\hat{\bm{F}}^{2D}$ and the 3D geometric features $\bm{F}^{3D}$. Specifically, the 2D prototypes $\bm{T}^{2D} \in \mathbb{R}^{\mathcal{C}^{\text{novel}} \times K}$ are built from $\hat{\bm{F}}^{2D}$ to avoid reliance on the VLMs during incremental stage, while the 3D prototypes $\bm{T}^{3D} \in \mathbb{R}^{\mathcal{C}^{\text{novel}} \times L}$ are obtained from proposal-based 3D features $\bm{F}^{3D}$.

\begin{algorithm}[t]
	\caption{Gated multimodal Prototype Imprinting.}
	\label{algo:gmproto}
	\begin{algorithmic}[1]
		\STATE \textbf{Input:} Aligned 2D features $\hat{\bm{F}}^{2D}$, 3D features $\bm{F}^{3D}$, prototypes $\bm{T}^{2D}$, $\bm{T}^{3D}$, momentum $\mu$;
		\STATE \textbf{Output:} updated prototypes $\bm{T}^{2D}$, $\bm{T}^{3D}$, fused classification scores $\bm{S}^{\text{fuse}}$.
		\vspace{4pt}
		\STATE {\color{gray}// Modality-specific Prototype Updating}
		\FOR{each novel class $c$}
		\STATE $\bar{\bm{F}}^{3D}_{c} = \operatorname{mean}(\bm{F}^{3D}_{c})$
		\STATE $\bm{T}^{3D}_{c} \leftarrow \mu \bm{T}^{3D}_{c} + (1 - \mu)\bar{\bm{F}}^{3D}_{c}$
		\STATE $\bar{\bm{F}}^{2D}_{c} = \operatorname{mean}(\hat{\bm{F}}^{2D}_{c})$
		\STATE $\bm{T}^{2D}_{c} \leftarrow \mu \bm{T}^{2D}_{c} + (1 - \mu)\bar{\bm{F}}^{2D}_{c}$
		\ENDFOR
		\vspace{4pt}
		\STATE {\color{gray}// Modality-specific Classification Scores}
		\STATE $\bm{S}^{3D} = \cos(\operatorname{norm}(\bm{F}^{3D}), \operatorname{norm}(\bm{T}^{3D}))$
		\STATE $\bm{S}^{2D} = \cos(\operatorname{norm}(\hat{\bm{F}}^{2D}), \operatorname{norm}(\bm{T}^{2D}))$
        \vspace{4pt}
		\vspace{2pt}
		\STATE {\color{gray}// Adaptive Multimodal Gating Fusion}
		\STATE $[\bm{\alpha}^{3D}, \bm{\alpha}^{2D}] = 
		\operatorname{Softmax}(\operatorname{MLP}([\bm{F}^{3D}; \hat{\bm{F}}^{2D}]))$
		\STATE $\bm{\gamma} = 
		\sigma(\operatorname{MLP}([\bm{F}^{3D}; \hat{\bm{F}}^{2D}]))$
		\STATE $\bm{S}^{\text{fuse}} =
		\bm{\gamma} \odot 
		(\bm{\alpha}^{3D} \odot \bm{S}^{3D} +
		\bm{\alpha}^{2D} \odot \bm{S}^{2D})$
		\vspace{2pt}
		\STATE \textbf{return} $\bm{T}^{2D}$, $\bm{T}^{3D}$, $\bm{S}^{\text{fuse}}$
	\end{algorithmic}
\end{algorithm}

Taking the 3D prototypes as an example, the model first determines the positive samples of novel-class objects based on the center-based label matching strategy~\cite{tr3d}. 
For each novel class $c$, the mean feature of its positive 3D samples in the current scene is denoted as $\bar{\bm{F}}_{c}^{3D}$. 
Since novel classes in incremental learning often have limited and unstable samples, directly updating the prototypes with such few-shot features may lead to feature overfitting. 
To address this, we introduce a momentum-based imprinting strategy that preserves historical information during prototype updating. 
The update rule is defined as:
\begin{equation}
	\bm{T}_{c}^{3D} \leftarrow 
	\mu\, \bm{T}_{c}^{3D} + (1 - \mu)\, \bar{\bm{F}}_{c}^{3D},
\end{equation}
where $\mu$ is the momentum coefficient controlling the update rate. 
Similarly, the 2D prototypes $\bm{T}_{c}^{2D}$ are updated using the aligned 2D features $\hat{\bm{F}}_{c}^{2D}$ in the same manner.
\begin{table*}[t]
	\caption{Batch incremental 3D object detection performance on ScanNet V2~\cite{scannet}. Results are reported under 1-way/9-way and 1-shot/5-shot settings. 
    Bold indicates the best performance, and underline indicates the second best.}
	\vspace{-5pt}
	\label{tab:main_scannet}
	\centering
	\setlength{\tabcolsep}{7pt}
	\resizebox{1\textwidth}{!}{
		\begin{tabular}{ccccccccccccc}
			\toprule
			\multirow{2.5}{*}{Methods}  & \multicolumn{3}{c}{1-way 1-shot} & \multicolumn{3}{c}{1-way 5-shot} & \multicolumn{3}{c}{9-way 1-shot} & \multicolumn{3}{c}{9-way 5-shot}  \\ 
            \cmidrule(lr){2-4} \cmidrule(lr){5-7} \cmidrule(lr){8-10} \cmidrule(lr){11-13}
			&Base & Novel & All & Base & Novel & All & Base & Novel & All& Base & Novel & All  \\
			
			\midrule
			
			Baseline & 71.47 & - & - & 71.47 & - & - & \textbf{72.77} & - & - & \textbf{72.77} & - & - \\
			Imprinting~\cite{imprinting} & 71.47 & 1.81 & 67.62 & 71.47 & 0.23 & 67.72 & 72.77 & 6.52 & 39.64 & 72.77 & 7.10 & 39.94 \\
			IL-DETR~\cite{incre_detr} & 69.78 & 0.03 & 65.91 & 65.63 & 0.35 & 62.00 & 65.77 & 6.02 & 35.90 & 67.05 & 13.82 & 40.43 \\
			SDCOT++~\cite{sdcot++} & 67.75 & 0.05 & 63.99 & 62.12 & 0.09 & 58.68 & 35.87 & 1.35 & 18.61 & 28.30 & 7.77 & 18.03 \\
			AIC3DOD~\cite{aic3dod} & 67.44 & 0.07 & 63.69 & 70.54 & 4.59 & 66.88 & 71.66 & 8.94 & 40.30 & 69.97 & 15.43 & 42.70 \\
			VLM-vanilla & \ul{71.81} & \ul{7.50} & \ul{68.24} & \ul{71.81} & \ul{14.09} & \ul{68.60} & 71.79 & \ul{17.12} & \ul{44.45} & 71.78 & \ul{16.72} & \ul{44.25} \\
			\rowcolor{red!5}
			FI3Det~(ours) & \textbf{72.85} & \textbf{35.58} & \textbf{70.78} & \textbf{72.84} & \textbf{38.48} & \textbf{70.94} & \ul{72.27} & \textbf{30.81} & \textbf{51.54} & \ul{72.28} & \textbf{30.23} & \textbf{51.26} \\
			
			\bottomrule
\end{tabular}}
	\vspace{-5pt}
\end{table*}

\begin{table*}[t]
	\caption{Batch incremental 3D object detection performance on SUN RGB-D~\cite{sunrgbd}. Results are reported under 1-way/5-way and 1-shot/5-shot settings. 
    Bold indicates the best performance, and underline indicates the second best.}
	\vspace{-5pt}
	\label{tab:main_sunrgbd}
	\centering
	\setlength{\tabcolsep}{7pt}
	\resizebox{1\textwidth}{!}{
		\begin{tabular}{ccccccccccccc}
			\toprule
			\multirow{2.5}{*}{Methods}  & \multicolumn{3}{c}{1-way 1-shot} & \multicolumn{3}{c}{1-way 5-shot} & \multicolumn{3}{c}{5-way 1-shot} & \multicolumn{3}{c}{5-way 5-shot}  \\ 
            \cmidrule(lr){2-4} \cmidrule(lr){5-7} \cmidrule(lr){8-10} \cmidrule(lr){11-13}

			& Base & Novel & All & Base & Novel & All & Base & Novel & All & Base & Novel & All \\
			\midrule
			Baseline & \ul{62.37} & - & - & \ul{62.37} & - & - & 61.58 & - & - & 61.58 & - & - \\
			Imprinting~\cite{imprinting} & 62.37 & 0.18 & 56.15 & 62.37 & 1.61 & 56.29 & 61.58 & 4.70 & 33.14 & 61.58 & 4.32 & 32.95 \\
			IL-DETR~\cite{incre_detr} & 61.81 & 0.05 & 55.63 & 61.72 & 0.02 & 55.55 & 58.27 & 0.74 & 29.50 & 58.90 & 0.25 & 29.57 \\
			SDCOT++~\cite{sdcot++} & 54.11 & 0.09 & 48.71 & 51.19 & 0.11 & 46.08 & 48.95 & 0.90 & 24.93 & 46.67 & 1.10 & 23.88 \\
			AIC3DOD~\cite{aic3dod} & 58.13 & 0.05 & 52.32 & 58.83 & 0.02 & 52.95 & 60.53 & 2.35 & 31.44 & 58.28  & 0.88 & 29.58 \\
			VLM-vanilla & 62.12 & \ul{5.72} & \ul{56.48} & 62.12 & \ul{11.93} & \ul{57.10} & \ul{62.10} & \ul{9.11} & \ul{35.60} & \ul{62.08} & \ul{10.22} & \ul{36.15} \\
			\rowcolor{red!5}
			FI3Det~(ours) & \textbf{63.06} & \textbf{67.29} & \textbf{63.48} & \textbf{63.05} & \textbf{73.17} & \textbf{64.07} & \textbf{62.49} & \textbf{15.27} & \textbf{38.88} & \textbf{62.49} & \textbf{26.81} & \textbf{44.65} \\
			\bottomrule
	\end{tabular}}
	\vspace{-10pt}
\end{table*}

\noindent\textbf{Multimodal Score Gating Fusion.} 
After obtaining the multimodal prototypes, we compute the cosine similarity between each scene’s features and the corresponding class prototypes to derive modality-specific classification scores.
Taking 3D features as an example, the class scores are computed as:
\begin{equation}
	\mathbf{S}^{3D} = 
	\cos\!\big(
	\operatorname{norm}(\bm{F}^{3D}),
	\operatorname{norm}(\bm{T}^{3D})
	\big),
\end{equation}
where $\mathbf{S}^{3D} \in \mathbb{R}^{N \times \mathcal{C}_{\text{novel}}}$ and the 2D scores $\mathbf{S}^{2D} \in \mathbb{R}^{N \times \mathcal{C}_{\text{novel}}}$ are computed in the same manner.

A straightforward fusion strategy is to directly sum the multimodal scores. However, this approach ignores the distinct characteristics of each modality, often leading to sub-optimal results. To overcome this limitation, we introduce a multimodal score gating fusion mechanism that adaptively combines 3D geometric and 2D semantic cues for improved object recognition. Specifically, adaptive gating functions are employed to learn the relative reliability of each modality and class:
\begin{equation}
	\begin{aligned}
		[\bm{\alpha}^{3D}; \bm{\alpha}^{2D}] 
		&= \operatorname{Softmax}\!\Big(
		\operatorname{MLP}\big(
		[\bm{F}^{3D}; \hat{\bm{F}}^{2D}]
		\big)
		\Big), \\
		\bm{\gamma} 
		&= \operatorname{Softmax}\!\Big(
		\operatorname{MLP}\big(
		[\bm{F}^{3D}; \hat{\bm{F}}^{2D}]
		\big)
		\Big),
	\end{aligned}
\end{equation}
where $\bm{\alpha}^{3D} \in \mathbb{R}^{N \times 1}$ and $\bm{\alpha}^{2D}\in \mathbb{R}^{N \times 1}$ control the modality-specific contributions, and $\bm{\gamma} \in \mathbb{R}^{N \times \mathcal{C}{\text{novel}}}$  re-balances class contributions to mitigate overconfident predictions from other classes. 
The fused classification scores $\bm{S}^{\text{fuse}} \in \mathbb{R}^{N \times \mathcal{C}{\text{novel}}}$ are then computed as:
\begin{equation}
	\bm{S}^{\text{fuse}} = 
	\bm{\gamma} \odot 
	\big(
	\bm{\alpha}^{3D} \odot \bm{S}^{3D} + 
	\bm{\alpha}^{2D} \odot \bm{S}^{2D}
	\big),
\end{equation}
where $\odot$ denotes element-wise multiplication.

\subsection{Total Loss Training}
\label{sec:ts}

\noindent\textbf{Base Training.} 
In the base training stage, we jointly optimize the detection loss and the auxiliary objectives for unknown objects. Both the base and pseudo boxes of unknown objects follow the same label assignment mechanism as in~\cite{tr3d, fcaf3d}. In addition, the 2D instance features adopt the same positive sample assignment as their corresponding pseudo 3D boxes to maintain spatial alignment.
The overall optimization objective is formulated as:
\begin{equation}
\begin{aligned}
\min_{\theta_{\text{base}}} \quad
& \mathcal{L}_{\text{det}}\big(\theta_{\text{base}}(\bm{\mathcal{X}}_{\text{base}}), \bm{\mathcal{Y}}_{\text{base}}\big) \\
& + \mathcal{L}_{\text{aux}}\big(\theta_{\text{base}}(\bm{\mathcal{X}}_{\text{base}}), \bm{\mathcal{Y}}_{\text{aux}}\big).
\end{aligned}
\end{equation}
Here, $\theta_{\text{base}}$ denotes the parameters of the base detector. 
$\mathcal{L}_{\text{det}}$ is the detection loss~\cite{tr3d, fcaf3d} including classification and box regression. The auxiliary loss  $\mathcal{L}_{\text{aux}}$ contains an objectness term $\mathcal{L}_{\text{aux-obj}}$, an unknown box regression term $\mathcal{L}_{\text{aux-box}}$, and a feature aligning term $\mathcal{L}_{\text{aux-feat}}$, jointly enhancing object-background separation, geometric alignment, and 2D-3D feature coherence.

\noindent\textbf{Incremental Learning.} 
In the incremental stage, the detector parameters are frozen, and only the prototypes and gating functions of novel classes are updated using few-shot samples. 
The objective function is defined as:
\begin{equation}
	\min_{\phi_{\text{new}}}
	\; \mathcal{L}_{\text{inc}}\big({\phi_{\text{new}}}(\bm{\mathcal{X}}_{\text{new}}), \bm{\mathcal{Y}}_{\text{new}}\big),
\end{equation}
where $\phi_\text{new}$ denotes the parameters of gating functions and $\mathcal{L}_{\text{inc}}$ is computed over novel classes. Please refer to the supplementary material for more detailed descriptions.

\section{Experiments}
\label{sec:experiments}

\subsection{Experimental Settings}\label{sec:data}
\noindent\textbf{Datasets and Setup.}
Since there is no existing dataset for 3D few-shot incremental object detection, we construct several few-shot incremental splits based on ScanNet V2~\cite{scannet} and SUN RGB-D~\cite{sunrgbd}. Following the class-splitting strategy in previous incremental works~\cite{sdcot, sdcot++, aic3dod}, we divide each dataset into base and novel classes in alphabetical order and build corresponding subsets. ScanNet V2 contains 1,201 training and 312 validation samples with 18 categories, while SUN RGB-D includes 5,285 training and 5,050 validation samples with 10 categories. 

Following previous incremental works~\cite{sdcot, sdcot++, aic3dod}, both datasets support batch and sequential few-shot incremental settings: in the batch setting, all novel classes are introduced simultaneously (\textit{i.e.}, ScanNet V2: 1/5-shot with 1-way and 9-way; SUN RGB-D: 1/5-shot with 1-way and 5-way), whereas in the sequential setting, novel classes are introduced across tasks (\textit{i.e.}, three 3-class tasks for ScanNet V2 and two tasks introducing 3 and 2 classes for SUN RGB-D, each with 5 samples per class). 

For the evaluation metric, we adopt mean Average Precision with an IoU threshold of 0.25 as the metric and report results separately for Base, Novel, and All categories.

\begin{table*}[t]
	\centering
	\caption{Sequential incremental results on ScanNet V2~\cite{scannet} and SUN RGB-D~\cite{sunrgbd}. Results are reported on 9-way 5-shot and 5-way 5-shot settings, respectively. 
    Bold indicates the best performance, and underline indicates the second best.
    }
	\label{tab:seq_scannet_sunrgbd}
	\vspace{-5pt}
	
	\setlength{\tabcolsep}{4pt} 
	
	\resizebox{\textwidth}{!}{
		\begin{tabular}{cccccccccccccccc}
			\toprule
			\multirow{3.5}{*}{Method} &
			\multicolumn{9}{c}{ScanNet V2}&
			\multicolumn{6}{c}{SUN RGB-D} \\
            \cmidrule(lr){2-10} \cmidrule(lr){11-16}
			& \multicolumn{3}{c}{Task1} & \multicolumn{3}{c}{Task2} & \multicolumn{3}{c}{Task3}
			& \multicolumn{3}{c}{Task1} & \multicolumn{3}{c}{Task2} \\
            \cmidrule(lr){2-4} \cmidrule(lr){5-7} \cmidrule(lr){8-10} \cmidrule(lr){11-13}  \cmidrule(lr){14-16}
			& Base & Novel & All & Base & Novel & All & Base & Novel & All
			& Base & Novel & All & Base & Novel & All \\
			\midrule
			Baseline & \textbf{72.77} & - & - & \textbf{72.77} & - & - & \textbf{72.77} & - & -
			& 61.58 & - & - & 61.58 & - & - \\
			Imprinting~\cite{imprinting} 
			& \ul{72.75} & 3.08 & \ul{55.35} & \ul{72.76} & 9.20 & \ul{47.34} & \ul{72.76} & 7.47 &  40.12
			& 62.23 & 4.37 & 40.53 & 61.62 & 5.85 & 33.74 \\
			IL-DETR~\cite{incre_detr} 
			& 62.53 & 6.43 & 48.51 & 36.18 &\ul{16.64}& 28.36 & 14.88 & 14.04  & 14.46
			& 58.53 & 0.07 & 36.61 & 53.50 & 0.40 & 26.95 \\
			SDCOT++~\cite{sdcot++} 
			& 43.38 & 0.82 & 32.75 & 15.11 & 10.24 & 13.17 & 7.61 & 0.59 & 4.10 
			& 47.30 & 0.16 & 29.62 & 22.43 & 0.03 & 11.23 \\
			AIC3DOD~\cite{aic3dod} & 70.72 & \ul{7.59} & 54.94 & 69.35 &11.99 & 46.40 & 66.88 & \ul{14.85} &   \ul{40.86} & 58.59 &  1.47 & 37.17 & 53.87   & 5.33 & 29.60 \\
			VLM-vanilla 
			& 71.79 & 2.39 & 54.44 & 71.77 & 9.36 & 46.81 & 71.78 & 8.80  &  40.29 
			&  \ul{62.86} & \ul{11.64} & \ul{43.66}  & \ul{62.08} & \ul{11.03} & \ul{36.55}  \\
			\rowcolor{red!5}
			FI3Det~(ours)	& 72.27 & \textbf{13.14} & \textbf{57.50} & 72.30 & \textbf{21.06} & \textbf{51.80} & 72.27 & \textbf{30.34} & \textbf{51.31}	& \textbf{63.56} & \textbf{13.02} & \textbf{44.61} & \textbf{62.49} & \textbf{19.04} & \textbf{40.76} \\
			\bottomrule
	\end{tabular}}
	\vspace{-10pt}
\end{table*}

\begin{table}[t]
	\centering
	\caption{Ablation study of key components including UOM, UOW, GPI on ScanNet V2.}
	\vspace{-5pt}
	\label{tab:components}
	\setlength{\tabcolsep}{8pt}
	\resizebox{1.0\linewidth}{!}{
		\begin{tabular}{ccccccc}
			\toprule
			No. & UOM & UOW & GPI & Base & Novel & All   \\ 
			\midrule
			1 & & &  & 71.81 & 14.09 & 68.60   \\
			2 & \checkmark &  &  & 72.73 & 25.43 & 70.10   \\
			3 & \checkmark & \checkmark &  & 72.83 & 32.46 & 70.61   \\
			4 & \checkmark &  & \checkmark & 72.73 & 28.94 & 70.30   \\
			\rowcolor{red!5}
			5 & \textbf{\checkmark} & \textbf{\checkmark} & \textbf{\checkmark} & 
			\textbf{72.84} & \textbf{38.48} & \textbf{70.94}  \\
			\bottomrule
		\end{tabular}
	}
	\vspace{-10pt}
\end{table}

\noindent\textbf{Implementation Details.}
We use TR3D~\cite{tr3d}  as the base detector for our experiments. All experiments are implemented using the mmdetection3d~\cite{mmdet3d} framework. In terms of unknown object learning, we first apply GroundingDINO~\cite{grounding} during base class training to generate 2D bounding boxes conditioned on textual prompts such as ``object''. Then, we employ a category-agnostic segmentation model~\cite{efficientsam} to obtain the corresponding 2D masks.
It is worth noting that, for the ScanNet V2 dataset, we follow the settings in ~\cite{spatiallm} and additionally provide background prompts. Besides, we set hyper-parameter $\sigma$ and $\mu$ to 0.5 and 0.999, respectively. 

\noindent\textbf{Baselines.}
We adopt two 2D few-shot incremental detection methods (Imprinting~\cite{imprinting}, IL-DETR~\cite{incre_detr}) and two state-of-the-art 3D incremental detectors (SDCOT++~\cite{sdcot++}, AIC3DOD~\cite{aic3dod}) as baselines. Imprinting uses prototypical weights for new classes, while IL-DETR improves novel-class generalization via self-supervised learning. For 3D detection, SDCOT++ leverages pseudo labels and distillation, and AIC3DOD further incorporates layout learning. Besides, following standard VLM-guided practice~\cite{mixsup, general}, we include a VLM-vanilla baseline using only pseudo 3D boxes without any of our proposed modules.

\subsection{Main Results}\label{sec:vssota}

\begin{figure*}[t] 
	\centering
	\includegraphics[width=1.0\textwidth]{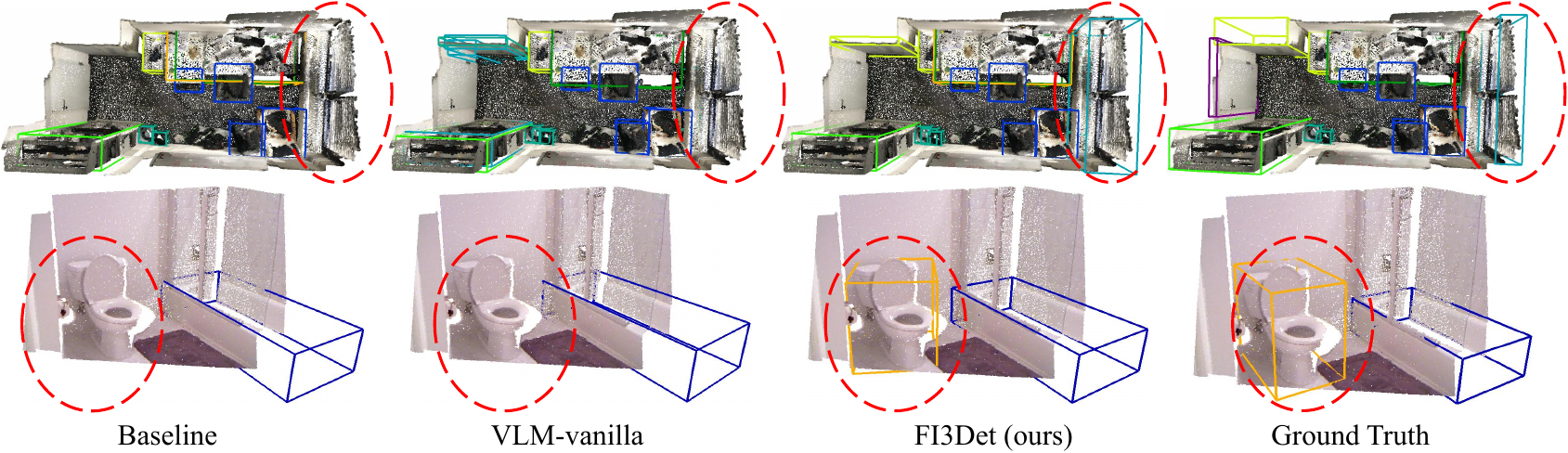}
	\vspace{-15pt}
	\caption{Qualitative comparison on the ScanNet V2~\cite{scannet}~(upper) and SUN RGB-D~\cite{sunrgbd}~(below). The red dashed circles highlight novel object categories are``\textit{\textcolor{cyan}{window}}'' and ``\textit{\textcolor{orange}{toilet}}''.}
	\label{fig:main_show}
	\vspace{-10pt}
\end{figure*}

\noindent\textbf{Batch Incremental Results.} In batch incremental settings, Tab.~\ref{tab:main_scannet} and Tab.~\ref{tab:main_sunrgbd} present the quantitative results on the ScanNet V2~\cite{scannet} and SUN RGB-D~\cite{sunrgbd} validation sets. 
We achieve state-of-the-art performance on novel classes. For example, on ScanNet V2, our method attains 38.48$\%$ mAP under the 1-way 5-shot setting, and on SUN RGB-D, it reaches 73.17$\%$ mAP under the 1-way 5-shot setting, demonstrating strong generalization to unseen categories. Furthermore, we also attain competitive performance on the base classes, which may be attributed to the richer supervision signals introduced in base training.

\noindent\textbf{Sequential Incremental Results.} 
Tab.~\ref{tab:seq_scannet_sunrgbd} reports the 5-shot sequential results on ScanNet V2 and SUN RGB-D. Our method achieves the best novel-class performance across all tasks. For instance, on ScanNet V2, we surpass the second-best method by 15.49$\%$ on Task 3, and on SUN RGB-D, we outperform it by 8.01$\%$ on Task 2. Despite the challenges of catastrophic forgetting in sequential incremental learning, our imprinting-based strategy effectively preserves base knowledge and ensures stable performance.

\noindent\textbf{Qualitative results.} We visualize the test results to demonstrate the reliability of our method on ScanNet V2~\cite{scannet} and SUN RGB-D~\cite{sunrgbd}. As shown in Fig.~\ref{fig:main_show}, our method successfully detects novel objects (\textit{e.g.}, ``\textit{\textcolor{cyan}{window}}'' and ``\textit{\textcolor{orange}{toilet}}'') while preserving accuracy on base categories.

\begin{figure}[t]
	\centering
	\includegraphics[width=1.0\linewidth]{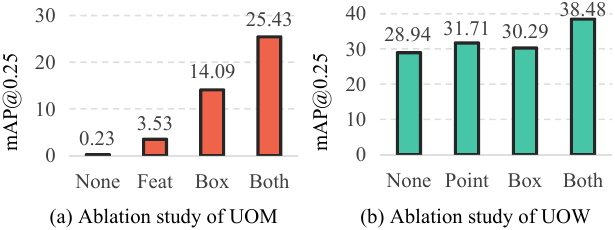}
	\vspace{-10pt}
	\caption{Ablation of different components in UOM and UOW.}
	\label{fig:vlm_unknown}
	\vspace{-15pt}
\end{figure}

\subsection{Ablation Study}\label{sec:ablation}
In this section, we investigate the role of each component under the 1-way 5-shot setting of ScanNet V2~\cite{scannet}.

\noindent\textbf{Effect of Different Components.} \label{sec:components} 
We first ablate the effects of different components of our model in Tab.~\ref{tab:components}. Variant 1 is a VLM-vanilla model based on TR3D~\cite{tr3d}, serving as the baseline. Variant 2 introduces unknown object mining (``UOM''), which combines 3D boxes with features and improves the novel class performance from 14.09$\%$ to 25.43$\%$. This shows that using features and 3D boxes in base training helps the model gain early awareness of novel objects. Adding unknown object weighting (``UOW'') in variant 3 further increases the novel class mAP to 32.46$\%$, showing that our weighting strategy suppresses noisy representations. When incorporating gated multimodal prototype imprinting (``GPI'') in the final version, the model achieves the best performance (mAP 38.48$\%$ for novel classes), demonstrating that fusing visual and geometric prototypes through adaptive gating enhances cross-modal knowledge transfer. Importantly, the base class accuracy remains stable across variants, suggesting that our design mitigates catastrophic forgetting. 

Moreover, we evaluate variant 4, which combines UOM and GPI, and observe a 3.51$\%$ gain, confirming the complementary nature of these components. We do not include a combination of UOW and GPI, as both modules rely on the representations established by UOM to function.

\noindent\textbf{Impact of Unknown Object Learning.}
We analyze the effectiveness of our unknown object learning by decomposing it into two modules: UOM and UOW.
As shown in Fig.~\ref{fig:vlm_unknown}, UOM consists of feature-level (``Feat'') and box-level (``Box'') mining, which individually improve novel-class mAP to 3.53$\%$ and 14.09$\%$, showing that both semantic alignment and spatial guidance aid novel recognition.
When combined, UOM achieves the highest 25.43$\%$ mAP, confirming their complementarity.
For UOW, point- and box-level weighting respectively enhance point reliability (31.71$\%$) and feature consistency inside boxes (30.29$\%$). The combination of both reaches 38.48$\%$ mAP, demonstrating that using joint weighting provides the most stable and effective unknown object learning.

\noindent\textbf{Analysis in Multimodal Imprinting.}\label{sec:mcir}
The modality-specific weights  $\bm{\alpha}^{*, *\in {\{3D}, {2D}\}}$ and fusion weight $\bm{\gamma}$ dynamically balance 3D and aligned features. 
Tab.~\ref{tab:gpi} shows that using only $\bm{\alpha}^{*}$ improves the novel class mAP from 32.46\% to 36.58\%, indicating that assigning modality weights can complement features. 
Using only $\bm{\gamma}$ also yields a gain, suggesting that it reduces the dominance of base classes and balances modality contributions. 
The combination of both methods leads to 38.48\% mAP on novel classes while maintaining stable base performance, demonstrating better transferability and robustness.

\begin{figure}[t]
	\centering
	\includegraphics[width=1.0\linewidth]{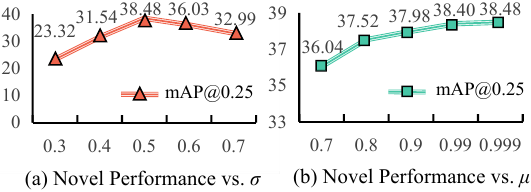}
	\vspace{-15pt}
	\caption{Performance of different hyper-parameters.}
	\label{fig:parameter}
    \vspace{-10pt}
\end{figure}

\noindent\textbf{Evaluation of Hyper-parameters.}
We analyze the influence of $\sigma$ and $\mu$ on novel classes in this section. 
As shown in Fig.~\ref{fig:parameter}(a), adjusting $\sigma$ controls the concentration of Gaussian weighting, and $\sigma{=}0.5$, which achieves 38.48$\%$ mAP for novel classes, shows an optimal trade-off. 
In Fig.~\ref{fig:parameter}(b), increasing $\mu$ consistently enhances performance, with the best result of mAP 38.48$\%$ obtained at $0.999$, indicating that a larger momentum helps stabilize training.

\begin{table}[t]
	\centering
	\caption{Ablation study of GPI components on ScanNet V2.}
	\vspace{-5pt}
	\label{tab:gpi}
	\setlength{\tabcolsep}{12pt}
	\resizebox{1\linewidth}{!}{
		\begin{tabular}{cccccc}
			\toprule
			No. & $\bm{\alpha}^{*}$ & $\bm{\gamma}$ & Base & Novel & All \\
			\midrule
			1 &  &  & 72.83 & 32.46 & 70.61 \\
			2 & \checkmark &  & 72.83 & 36.58 & 70.84 \\
			3 &  & \checkmark & \textbf{72.87} & 34.68 & 70.73 \\
			\rowcolor{red!5}
			4 & \textbf{\checkmark} & \textbf{\checkmark} & 
			72.84 & \textbf{38.48} & \textbf{70.94} \\
			\bottomrule
		\end{tabular}
	}
\end{table}

\section{Conclusion}\label{sec:conclusion}
In this paper, we propose a novel framework for few-shot incremental 3D object detection that learns new categories from limited annotations without forgetting previous ones. In the base stage, a VLM-guided module leverages VLM priors to discover unseen objects and refine their 2D semantics and class-agnostic 3D representations through point and box weights. In the incremental stage, a gated multimodal prototype module aligns 2D semantics and 3D geometry to construct prototypes for novel objects. Experiments on ScanNet V2 and SUN RGB-D demonstrate the superior performance and generalization ability of our approach.

\section*{Acknowledgments} This work was supported by the National Key R\&D Program of China No. 2024YFC3015801, National Science Fund of China (Grant Nos. 62361166670, 62276144, U24A20330) and Basic Research Program of Jiangsu under Grant No. BK20253028. This work was also supported by the Ministry of Education, Singapore, under its MOE Academic Research Fund Tier 2 (MOE-T2EP20124-0013).

\appendix

\section{Overview}
\label{sec:intor}
In this supplementary material, we first provide a detailed description of our training details (§~\ref{sec:detail}), including base training details~(§~\ref{sec:base}) and incremental learning details~(§~\ref{sec:incre}).
We then describe the detailed category split information of ScanNet V2 and SUN RGB-D (§~\ref{sec:data_split}).
Next, we present additional experiments (§~\ref{sec:more_exp}), including  the performance of unknown object learning on fully incremental 3D object detection (§~\ref{sec:full_incr}), the performance of our method when adopting different VLM-based models (§~\ref{sec:others}), more qualitative visualizations (§~\ref{sec:visual}), and the results under alternative category split settings (§~\ref{sec:counts}).
Finally, we discuss the limitations and future work (§~\ref{sec:limitations}) of our proposed approach.

\section{Training Details}\label{sec:detail}  
\subsection{Base Training Details}\label{sec:base}  

In this section, we provide a detailed description of how our framework leverages unknown objects during the base class training stage. Although these objects do not have semantic labels and therefore cannot contribute to the classification loss, their location and feature representation provide valuable cues to enhance the model’s generalization ability to novel categories. We incorporate unknown objects through three auxiliary supervisory signals:  foreground supervision, feature supervision, and regression supervision. 

\noindent\textbf{Foreground Supervision.}
Given a pseudo box $\mathcal{B}_j$, we assign each point $p_e$ a continuous foreground score $w_e \in [0,1]$ with $\operatorname{Sigmoid}$ function if it falls inside this pseudo box.
This target score $w_e=w_j^{\text{box}} w_{e,j}^{\text{point}}$ is determined by its spatial proximity to the box center and the internal feature consistency of the box, as described in the main paper. 
Let $o_e$ denote the predicted objectness probability. We supervise the objectness branch exclusively on unknown-object regions using a combination of binary cross-entropy and Dice loss~\cite{diceloss}:
\begin{equation}
\mathcal{L}_{\text{obj}}
=
\operatorname{BCE}(o_e, w_e)
+
\operatorname{Dice}(o_e, w_e).
\end{equation}

This supervision is independent of semantic categories and therefore naturally supports incremental learning, enabling the model to acquire foreground awareness for unseen objects even without manual labels.

\noindent\textbf{Feature Supervision.}
To further enhance semantic understanding of unknown objects, we introduce feature supervision for points inside pseudo boxes. As mentioned in the main paper, each unknown object has a pseudo 3D box $\mathcal{B}_j$ and an instance feature $\mathcal{F}_j^{2D}$.
For all points falling inside $\mathcal{B}_j$, we enforce cosine directional alignment:

\begin{equation}
\mathcal{L}_{\text{feat}}=
\frac{1}{Z_j}
\sum_{p_e \in \mathcal{B}_j}
\big(1 - \cos(\hat{f_e^{2D}}, f_j^{2D})\big)\, w_e,
\end{equation}

where $w_e$ is the soft weight / target score and $\cos$ is the cosine similarity, which measures the directional alignment between two feature vectors. This loss encourages the internal points of a pseudo box to form consistent semantic embeddings, allowing the model to inherit semantic priors for novel classes during the base training.

\noindent\textbf{Regression Supervision.}
Beyond the above parts, we also guide the model to learn the geometric structure of unknown objects. Since the regression branch does not involve any semantic information, we directly share it with the base class regression head, enabling the geometric priors learned by the network to naturally generalize to unknown categories. For point $p_e$ inside pseudo box $\mathcal{B}_j$, let $\hat{r}_e$ denote the predicted bounding box and $r_e^{*}$ the geometric parameters of the pseudo 3D box. 
We apply a soft-weighted regression loss with DIOU loss~\cite{diou}:
\begin{equation}
\mathcal{L}_{\text{reg}}^{\text{unk}}
=\frac{1}{Z_j}
\sum_{p_e \in \mathcal{B}_j}
\big(1 - \operatorname{DIOU}
(\hat{r}_e, r_j^{*})\big)\, w_e.
\end{equation}
This formulation provides the model with localization cues for unknown objects during base training, enabling it to acquire spatial awareness of novel categories. 

\textbf{Overall.} By combining the above objectives, we obtain the total auxiliary loss associated with unknown objects:
\begin{equation}
\mathcal{L}_{\text{aux}}
=\mathcal{L}_{\text{obj}}
+ \mathcal{L}_{\text{feat}}
+ \mathcal{L}_{\text{reg}}^{\text{unk}}.
\end{equation}

\subsection{Incremental Learning Details}\label{sec:incre}  
As introduced in the main paper, during the incremental learning stage, the fusion weights $\alpha^{\text{3D}} \in \mathbb{R}^{N \times 1}$ and $\alpha^{\text{2D}} \in \mathbb{R}^{N \times 1}$ control the modality-specific contributions of the 3D and 2D branches, while $\gamma \in \mathbb{R}^{N \times C_{\text{novel}}}$ re-balances per-class responses to mitigate overconfident predictions from other categories. To update these weighting parameters, we introduce an incremental loss $\mathcal{L}_{\text{inc}}$. Let $s^{\text{fuse}}_e$ denote the fused prediction after applying the modality- and class-wise weights, and let $y_e$ be the one-hot target for the novel categories. The incremental loss is defined using a simple positive-negative supervision:
\begin{equation}
\mathcal{L}_{\text{inc}}
=
(1 - s^{\text{fuse}}_e)\, y_e
+
s^{\text{fuse}}_e(1 - y_e).
\end{equation}

This supervision enables the model to gradually form discriminative boundaries for novel categories.

\noindent\section{Dataset Split Details}\label{sec:data_split}  
In this section, we provide more category split information for ScanNet V2~\cite{scannet} and SUN RGB-D~\cite{sunrgbd} mentioned in the main paper, including both batch-incremental and sequence-incremental settings.

\textbf{ScanNet V2}~\cite{scannet} contains 1,201 training samples and 312 validation samples, annotated with 18 object categories: 
\emph{bathtub, bed, bookshelf, cabinet, chair, counter, curtain, desk, door, garbagebin, picture, refrigerator, showercurtain, sink, sofa, table, toilet, and window}  in alphabetical order. For batch incremental settings, we design four few-shot incremental detection settings to evaluate the model’s generalization ability:
\begin{itemize}
	\item \textbf{1-way 1-shot:} 17 base classes (\emph{\textcolor{MidnightBlue}{bathtub–toilet}}), 
	1 novel class (\emph{\textcolor{BrickRed}{window}}) with 1 labeled sample.
	
	\item \textbf{1-way 5-shot:} Same as above, but with 5 labeled samples for the novel class.
	
	\item \textbf{9-way 1-shot:} 9 base classes (\emph{\textcolor{MidnightBlue}{bathtub–door}}), 
	9 novel classes (\emph{\textcolor{BrickRed}{garbagebin–window}}) with 1 labeled sample per novel class.
	
	\item \textbf{9-way 5-shot:} Same as above, but with 5 labeled samples per novel class.
\end{itemize}

For the sequential incremental setting, we initialize the model with 9 base classes (\emph{\textcolor{MidnightBlue}{bathtub–door}}) and introduce 3 novel classes at each incremental step, each with 5 labeled samples per novel class, resulting in three tasks, namely:

\begin{itemize}
	\item \textbf{Task 1}: \emph{\textcolor{BrickRed}{garbagebin, picture, refrigerator}}. 
	\item \textbf{Task 2}: \emph{\textcolor{BrickRed}{showercurtain, sink, sofa}}.
	\item \textbf{Task 3}: \emph{\textcolor{BrickRed}{table, toilet, window}}.
\end{itemize}

\textbf{SUN RGB-D}~\cite{sunrgbd} consists of 5,285 training samples and 5,050 validation samples, annotated with 10 object categories: 
\emph{bathtub, bed, bookshelf, chair, desk, dresser, night\_stand, sofa, table, and toilet} in alphabetical order.
Similar to ScanNet V2, we design four few-shot batch incremental detection settings to evaluate the model:
\begin{itemize}
	\item \textbf{1-way 1-shot:} 9 base classes (\emph{\textcolor{MidnightBlue}{bathtub–table}}), 1 novel class (\emph{\textcolor{BrickRed}{toilet}}) with 1 labeled sample.
	\item \textbf{1-way 5-shot:} Same as above, but with 5 labeled samples for the novel class.
	\item \textbf{5-way 1-shot:} 5 base classes (\emph{\textcolor{MidnightBlue}{bathtub–desk}}), 5 novel classes (\emph{\textcolor{BrickRed}{dresser–toilet}}) with 1 labeled sample per novel class.
	\item \textbf{5-way 5-shot:} Same as above, but with 5 labeled samples per novel class.
\end{itemize}

For the sequential incremental setting, we initialize the model with 5 base classes (\emph{\textcolor{MidnightBlue}{bathtub–desk}}) and introduce 3 novel classes an 2 novel classes sequentially, each with 5 labeled samples per novel class, resulting in two tasks:

\begin{itemize}
	\item \textbf{Task 1}: \emph{\textcolor{BrickRed}{dresser, night\_stand, sofa}}. 
	\item \textbf{Task 2}: \emph{\textcolor{BrickRed}{table, toilet}}.

\end{itemize}

\section{More Experiments} \label{sec:more_exp}

\subsection{Results on Fully Incremental Settings} \label{sec:full_incr}
We extend our VLM-guided Unknown object learning (denoted as UOL) to the fully incremental setting, where the model has access to all annotated novel categories during the incremental stage. To ensure a clean evaluation, we implement a simplified version based on TR3D that retains only the pseudo labeling mechanism (denoted as Ours$^*$) to highlight the contribution of our base training strategy.

As shown in Tab. \ref{tab:full_incre}, when integrating our proposed VLM-guided base training strategy (denoted as Ours$^*$ + UOL) into Ours$^*$, the model achieves consistent and stable improvements across the Base, Novel, and All metrics under both the 1-way and 9-way incremental configurations. In the 1-way setting, the Novel mAP increases from 55.52 to 59.76, and the All mAP improves from 71.75 to 73.45. In the 9-way setting, the Novel mAP similarly rises from 69.63 to 71.91, and the All mAP improves from 72.12 to 73.66. These results demonstrate the generalization ability of our base training strategy in fully incremental scenarios.

\begin{table}[!t]
	\caption{Batch incremental 3D object detection results with fully labeled novel objects on ScanNet V2. All models are based on the TR3D baseline. Results are reported under 1-way and 9-way configurations. ``Base'' denotes base classes, ``Novel'' denotes novel classes, and ``All'' indicates the overall mean AP@0.25. }
	\label{tab:full_incre}
	\centering
	\resizebox{1.0\linewidth}{!}{
		\begin{tabular}{ccccccc}
			\toprule
			\multirow{4}{*}{Method}  

			& \multicolumn{3}{c}{1-way} & \multicolumn{3}{c}{9-way} \\ 
			\cmidrule(r){2-4} \cmidrule(r){5-7} 
			& Base & Novel & All & Base & Novel & All  \\
			\midrule
			Baseline & 71.47 & - & - & 72.77 & - & - \\
            Ours$^*$~\cite{sdcot} 
			& 72.71 & 55.52 & 71.75 & 74.61 & 69.63 & 72.12 \\
			\rowcolor{red!10}
			Ours$^*$+UOL
			& \textbf{74.26} & \textbf{59.76} & \textbf{73.45} & \textbf{75.40} & \textbf{71.91} & \textbf{73.66} \\
			\bottomrule
	\end{tabular}}
\end{table}

\subsection{Results based on Other VLMs} \label{sec:others}
In this experiment, we replace the VLM backbone to further validate the generalization of our framework shown in Tab.~\ref{tab:vlm_results}. Specifically, we substitute GroundingDINO~\cite{grounding} with YOLO-World~\cite{yolo_world}, which also supports open-vocabulary detection but requires explicit category prompts to perform inference. To ensure a comparison, we provide YOLO-World with a comprehensive prompt set containing 50 common indoor categories, including:

\emph{``chair'', ``table'', ``sofa", ``bed'', ``desk'', ``cabinet'', ``shelf'', ``lamp'', ``door'', ``window'', ``television'', ``refrigerator'', ``washing machine'', ``microwave'', ``fan'', ``air conditioner'', ``sink'', ``toilet'', ``bathtub'', ``shower'', ``mirror'', ``carpet'', ``pillow'', ``blanket'', ``curtain'', ``picture'', ``vase'', ``clock'', ``books'', ``laptop'', ``keyboard'', ``mouse'', ``monitor'', ``printer'', ``trash bin'', ``cup'', ``plate'', ``bottle'', ``kettle'', ``knife'', ``wardrobe'', ``shoe'', ``bag'', ``clothes'', ``towel'', ``plant'', ``cushion'', ``stool'', ``nightstand'', and ``drawer''}.

Moreover, to preserve the incremental learning protocol, no category information produced by the VLM is retained during base training. As shown in Tab. \ref{tab:vlm_results}, both VLM-based variants outperform the baseline by a large margin. Among them, GroundingDINO achieves the best overall performance on both ScanNet V2 and SUN RGB-D, particularly in novel class detection (\textit{e.g.}, +3.85 and +2.84 mAP over YOLO-World under 9-way and 5-way 5-shot settings, respectively). These findings demonstrate that our proposed framework can flexibly integrate different VLMs while maintaining strong incremental detection capability.

\subsection{More Quantitative Results} \label{sec:visual}
\begin{figure*}[t]
	\centering
	\includegraphics[width=1.0\textwidth]{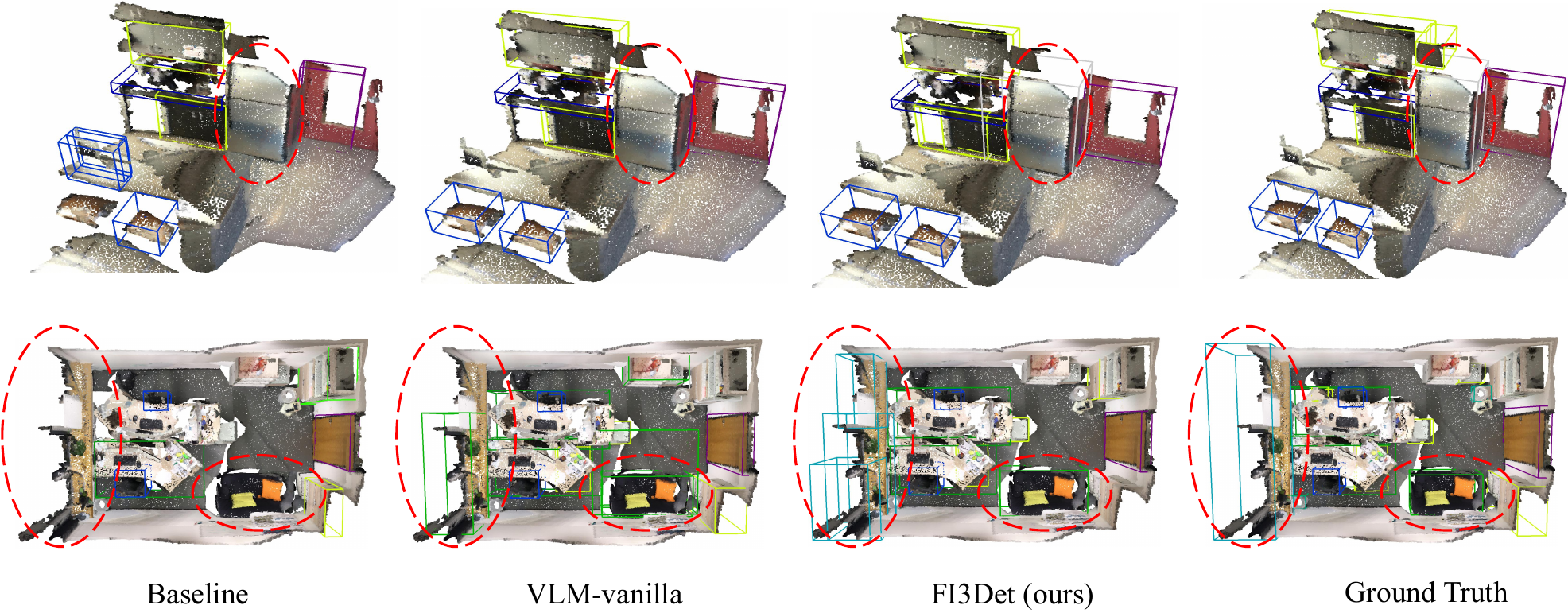}
	\caption{Qualitative comparison on the ScanNet V2~\cite{sunrgbd}. The red dashed circles highlight novel object categories ``\textit{\textcolor{ForestGreen}{sofa}}'', ``\textit{\textcolor{gray}{refrigerator}}'', and ``\textit{\textcolor{cyan}{window}}''. }
	\label{fig:scan_show}
\end{figure*}

\begin{figure*}
	\centering
	\includegraphics[width=1.0\textwidth]{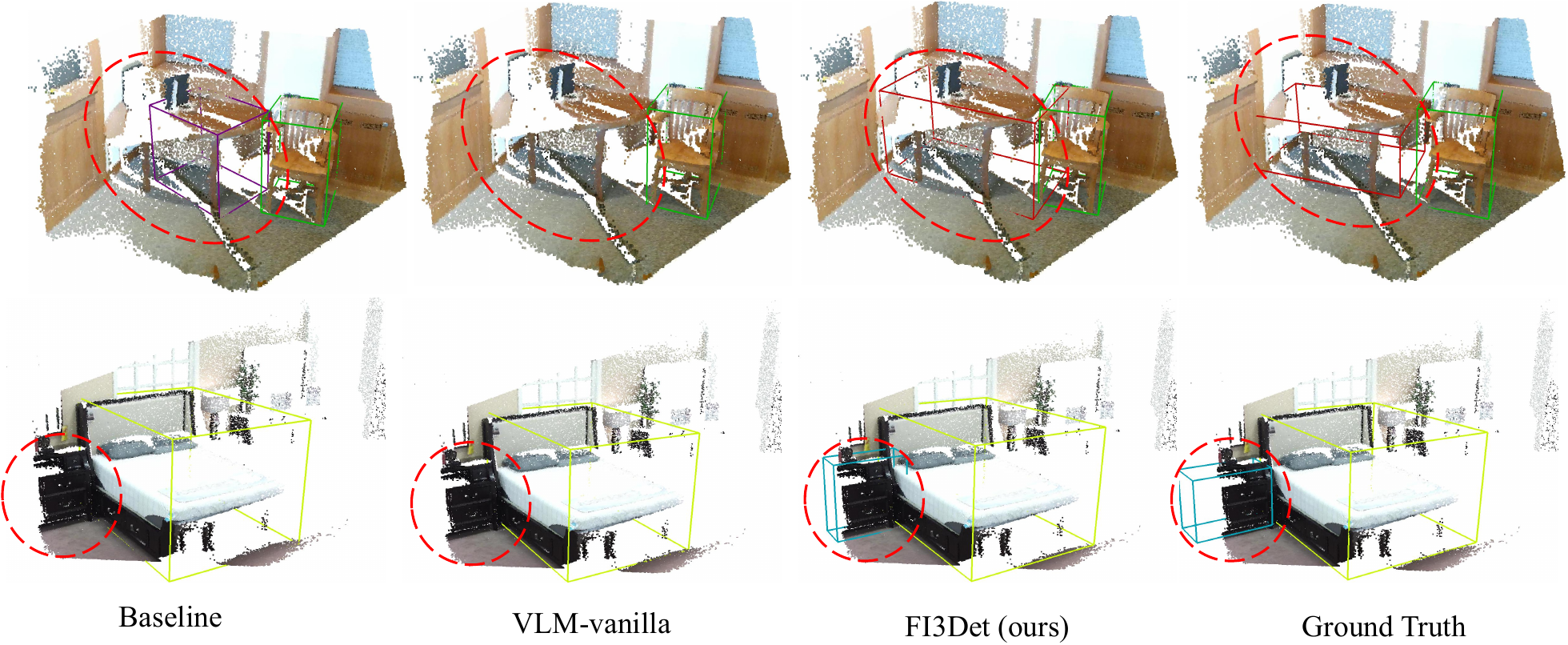}
	\caption{Qualitative comparison on the SUN RGB-D~\cite{sunrgbd}. The red dashed circles highlight novel object categories``\textit{\textcolor{BrickRed}{table}}'' and ``\textit{\textcolor{cyan}{dresser}}''. }
	\label{fig:sun_show}
\end{figure*}

In this section, we provide additional quantitative results on ScanNet V2~\cite{scannet} and SUN RGB-D~\cite{sunrgbd}. The predicted bounding boxes on these two datasets are shown in Fig.~\ref{fig:scan_show} and Fig.~\ref{fig:sun_show}. In the qualitative results, the red dashed circles highlight novel object categories, including ``\textit{\textcolor{ForestGreen}{sofa}}'', ``\textit{\textcolor{gray}{refrigerator}}'', and ``\textit{\textcolor{cyan}{window}}'' in ScanNet V2, as well as ``\textit{\textcolor{BrickRed}{table}}'' and ``\textit{\textcolor{cyan}{dresser}}'' in SUN RGB-D. Compared with Baseline and VLM-vanilla, which often miss or inaccurately localize these novel categories, our FI3Det produces more accurate and stable detections that closely match the ground truth, demonstrating stronger generalization to novel classes under few-shot incremental settings.

\subsection{Results on Alternative Category Splits} \label{sec:counts}
\begin{figure}
	\centering
	\includegraphics[width=1.0\linewidth]{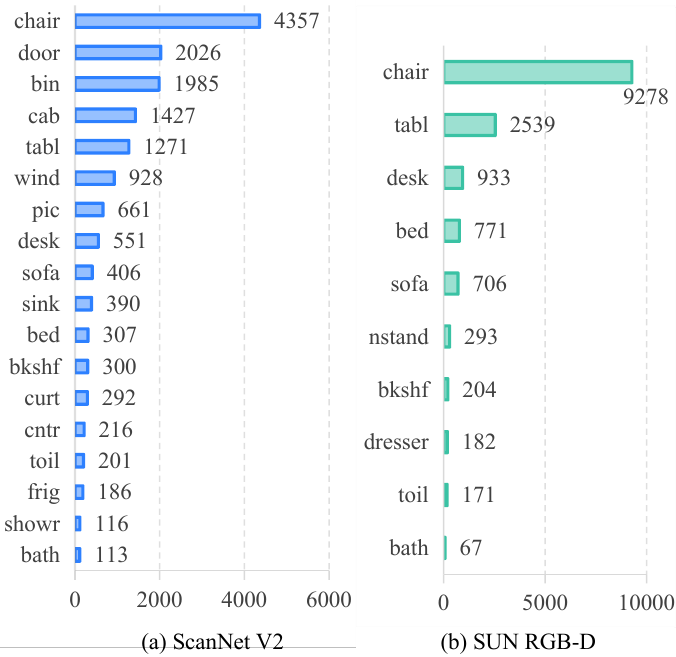}
	\caption{Statistical analysis of the number of instances for each category in ScanNet V2 and SUN RGB-D.}
	\label{fig:count}
\end{figure}

In the main paper, we follow the setting of~\cite{sdcot++}, where novel categories for few-shot incremental 3D object detection are selected based on alphabetical order. In this section, we further explore alternative category split strategies to verify the robustness of our method. As shown in Fig.~\ref{fig:count}, both ScanNet v2 and SUN RGB-D datasets contain a large number of object instances, but their category distributions are highly imbalanced, exhibiting a clear long-tailed property. Rather than randomly selecting novel categories, which can bias the evaluation, we divide the base and novel categories based on the number of instances per class.
We design four incremental detection settings to evaluate the model’s generalization ability based on ScanNet V2~\cite{scannet} and SUN RGB-D~\cite{sunrgbd}:
\begin{itemize}
	
	\item \textbf{9-way 1-shot:} 9 base classes (\emph{\textcolor{MidnightBlue}{chair–sofa}}), 
	9 novel classes (\emph{\textcolor{BrickRed}{sink–bathtub}}) with 1 labeled sample per novel class on ScanNet V2.
	
	\item \textbf{9-way 5-shot:} Same as above, but with 5 labeled samples per novel class on ScanNet V2.
	
	\item \textbf{5-way 1-shot:} 5 base classes (\emph{\textcolor{MidnightBlue}{chair–sofa}}), 5 novel classes (\emph{\textcolor{BrickRed}{night\_stand–bathtub}}) with 1 labeled sample per novel class on SUN RGB-D.
	\item \textbf{5-way 5-shot:} Same as above, but with 5 labeled samples per novel class on SUN RGB-D.
\end{itemize}

Tab.~\ref{tab:main_combined} presents the results on ScanNet V2 and SUN RGB-D. For ScanNet V2, results are reported under 9-way 1-shot and 9-way 5-shot configurations, while for SUN RGB-D, results are provided under 5-way 1-shot and 5-way 5-shot configurations. As shown in the table, our proposed FI3Det achieves consistently superior performance across all configurations. In particular, it significantly improves detection on novel categories while maintaining strong performance on base categories.

\section{Limitations} \label{sec:limitations}
This work leverages vision-language models (VLMs) to learn general semantic representations during the base class training stage, enabling the detector to achieve adaptability when encountering novel objects. Although we assume that the robot has a basic exploration of the environment before task switching, this setting is reasonable in most indoor scenarios (\eg, homes or offices) but may present limitations in more complex or dynamic environments.

In the future, we plan to enhance the robustness of the network through improved architectural designs, enabling more stable learning in real-world embodied perception tasks. Moreover, although our method is capable of handling indoor environments, large-scale outdoor autonomous driving scenarios remain a relatively underexplored domain, which we plan to investigate further in our future work.

\begin{table*}
	\caption{Batch few-shot incremental 3D object detection results with different Vision-Language Models (VLMs) on ScanNet V2 and SUN RGB-D. 
		Results are reported under 9-way/5-way and 1-shot/5-shot configurations. 
		``Base'' denotes base classes, ``Novel'' denotes novel classes, and ``All'' indicates the overall mean AP@0.25.}
	\label{tab:vlm_results}
	\centering
	\resizebox{1.0\textwidth}{!}{
		\begin{tabular}{ccccccccccccc}
			\toprule
			\multirow{4}{*}{Method}  
			& \multicolumn{6}{c}{ScanNet V2} 
			& \multicolumn{6}{c}{SUN RGB-D} \\ 
            \cmidrule(r){2-7} \cmidrule(r){8-13} 
			& \multicolumn{3}{c}{9-way 1-shot} & \multicolumn{3}{c}{9-way 5-shot} 
			& \multicolumn{3}{c}{5-way 1-shot} & \multicolumn{3}{c}{5-way 5-shot} \\ 
			\cmidrule(r){2-4} \cmidrule(r){5-7} \cmidrule(r){8-10} \cmidrule(r){11-13}
			& Base & Novel & All & Base & Novel & All & Base & Novel & All & Base & Novel & All \\
			\midrule
			Baseline 
			& \textbf{72.77} & 6.52 & 39.64 & \textbf{72.77} & 7.10 & 39.94 
			& 61.58 & 4.70 & 33.14 & 61.58 & 4.32 & 32.95 \\
			YOLO-World~\cite{yolo_world} 
			& 72.43 & 28.09 & 50.76 & 72.44 & 26.38 & 49.91 
			& 61.77 & 16.24 & 39.01 & 61.77 & 23.97 & 42.87 \\
			\rowcolor{red!10}
			GroundDINO~(ours)~\cite{grounding} 
			& 72.27 & \textbf{30.81} & \textbf{51.54} & 72.28 & \textbf{30.23} & \textbf{51.26} 
			& \textbf{62.49} & \textbf{15.27} & \textbf{38.88} & \textbf{62.49} & \textbf{26.81} & \textbf{44.65} \\
			\bottomrule
	\end{tabular}}
\end{table*}

\begin{table*}[t]
	\caption{Batch few-shot incremental 3D object detection performance on ScanNet V2~\cite{scannet} and SUN RGB-D~\cite{sunrgbd}. Results are reported under 9-way/5-way and 1-shot/5-shot configurations. ``Base'' denotes base classes, ``Novel'' denotes novel classes, and ``All'' indicates the overall mean AP@0.25. The base and novel categories are divided according to the number of object instances in each class.}
	\label{tab:main_combined}
    \setlength{\tabcolsep}{7pt}
	\centering
	\resizebox{1.0\textwidth}{!}{
		\begin{tabular}{ccccccccccccc}
			\toprule
			\multirow{4}{*}{Methods}  
			& \multicolumn{6}{c}{ScanNet V2} 
			& \multicolumn{6}{c}{SUN RGB-D} \\ 
            \cmidrule(r){2-7} \cmidrule(r){8-13} 
			& \multicolumn{3}{c}{9-way 1-shot} & \multicolumn{3}{c}{9-way 5-shot} 
			& \multicolumn{3}{c}{5-way 1-shot} & \multicolumn{3}{c}{5-way 5-shot} \\ 
			\cmidrule(r){2-4} \cmidrule(r){5-7} \cmidrule(r){8-10} \cmidrule(r){11-13}
			& Base & Novel & All & Base & Novel & All & Base & Novel & All & Base & Novel & All \\
			\midrule
			Imprinting~\cite{imprinting} 
			& 66.84 & 4.33 & 38.90 & 66.84 & 10.94 & 38.90 
			& 66.69 & 0.86 & 33.77 & 66.69 & 0.62 & 31.10 \\
			IL-DETR~\cite{incre_detr} 
			& 61.36 & 6.50 & 33.93 & 57.64 & 20.12 & 38.88 
			& 65.65 & 0.09 & 32.87 & 63.36 & 0.19 & 31.77 \\
			SDCOT++~\cite{sdcot++} 
			& 48.26 & 5.38 & 23.82 & 27.38 & 19.41 & 23.40 
			& 60.86 & 0.02 & 30.44 & 51.51 & 0.06 & 25.78 \\
			AIC3DOD~\cite{aic3dod} 
			& 66.82 & 5.14 & 35.98 & 66.97 & 10.09 & 38.53 
			& 66.72 & 0.05 & 33.39 & 66.51 & 0.07 & 33.29 \\
			VLM-vanilla 
			& 67.58 & 9.06 & 44.61 & 67.56 & 21.65 & 44.61 
			& 67.19 & 4.18 & 35.67 & 67.19 & 3.75 & 35.47 \\
			\rowcolor{red!10}
			FI3Det~(ours) 
			& \textbf{67.59} & \textbf{24.18} & \textbf{45.89} 
			& \textbf{67.63} & \textbf{38.63} & \textbf{53.10} 
			& \textbf{68.15} & \textbf{8.92} & \textbf{38.54} 
			& \textbf{68.16} & \textbf{20.46} & \textbf{44.31} \\
			\bottomrule
	\end{tabular}}
\end{table*}
\clearpage

\clearpage
{
    \small
    \bibliographystyle{ieeenat_fullname}
    \bibliography{main}
}

\end{document}